\documentclass[letterpaper]{article} 
\usepackage{aaai2026}  
\usepackage{times}  
\usepackage{helvet}  
\usepackage{courier}  
\usepackage[hyphens]{url}  
\usepackage{graphicx} 
\usepackage{natbib}  
\usepackage{caption} 
\usepackage{algorithm}

\usepackage{newfloat}
\usepackage{listings}
\DeclareCaptionStyle{ruled}{labelfont=normalfont,labelsep=colon,strut=off} 
\floatstyle{ruled}
\newfloat{listing}{tb}{lst}{}
\floatname{listing}{Listing}
\usepackage{amssymb,amsmath,latexsym}
\usepackage{subfig}
\usepackage[noend]{algpseudocode}
\usepackage{paralist}
\usepackage{xurl}
\usepackage{xspace}
\usepackage{xcolor}

\newcommand{\Prob}{\ensuremath{{Pr}}}
\newcommand{\ProbSafe}{\ensuremath{\Prob_{\sf safe}}}

\newcommand{\pos}{\ensuremath{p}}

\newcommand{\traj}{\ensuremath{{\sf traj}}}

\newcommand{\action}{\ensuremath{a}}
\newcommand{\plan}{\ensuremath{\pi}}
\newcommand{\PlanSet}{\ensuremath{\Pi}}

\newcommand{\formation}{\ensuremath{F}}

\newcommand{\FirstTime}{\ensuremath{t_{\sf first}}}
\newcommand{\StartTime}{\ensuremath{t_{0}}}
\newcommand{\EndTime}{\ensuremath{t_{\sf end}}}
\newcommand{\LastTime}{\ensuremath{t_{\sf last}}}

\newcommand{\drone}{\ensuremath{\upsilon}}
\newcommand{\DroneSet}{\ensuremath{\mathcal{V}}}

\newcommand{\HiddenDroneSet}{\ensuremath{\DroneSet_{\sf hidden}}}
\newcommand{\FallenDroneSet}{\ensuremath{\DroneSet_{\sf fallen}}}
\newcommand{\ActiveDroneSet}{\ensuremath{\DroneSet_{\sf active}}}
\newcommand{\EvacuateDroneSet}{\ensuremath{\DroneSet_{\sf evacuate}}}

\newcommand{\cell}{\ensuremath{c}}
\newcommand{\tile}{\ensuremath{\tau}}
\newcommand{\TileSet}{\ensuremath{\mathcal{T}}}

\newcommand{\StageSpace}{\ensuremath{\mathcal{Z}^{\sf stage}}}
\newcommand{\StageZone}{\ensuremath{\mathcal{H}^{\sf stage}}}
\newcommand{\FallZone}{\ensuremath{\mathcal{H}}}
\newcommand{\HitZone}{\ensuremath{\mathcal{H}^{\sf hit}}}
\newcommand{\SafeZone}{\ensuremath{\mathcal{H}^{\sf safe}}}
\newcommand{\ParkSpace}{\ensuremath{\mathcal{Z}^{\sf park}}}

\newcommand{\score}{{\sf score}}

\newtheorem{defn}{Definition}          

\newcommand{\commentc}[1]{}
\newcommand{\commentm}[1]{}

\renewcommand{\commentc}[1]{{\color{blue} **Chiu: #1**}}
\renewcommand{\commentm}[1]{{\color{orange} **Minhyuk: #1**}}

\title{Robust Evacuation for Multi-Drone Failure in Drone Light Shows}
\author{
    Minhyuk Park\textsuperscript{\rm 1},
    Aloysius K. Mok\textsuperscript{\rm 2},
    Tsz-Chiu Au\textsuperscript{\rm 1}
}
\affiliations{
    \textsuperscript{\rm 1}Department of Computer Science, Texas State University, USA\\
    \textsuperscript{\rm 2}Department of Computer Science, The University of Texas at Austin, USA\\
    dmz44@txstate.edu, mok@cs.utexas.edu, chiu.au@txstate.edu
}

\begin{document}

\maketitle

\begin{abstract}

Drone light shows have emerged as a popular form of entertainment in recent years. However, several high-profile incidents involving large-scale drone failures---where multiple drones simultaneously fall from the sky---have raised safety and reliability concerns. To ensure robustness, we propose a drone parking algorithm designed specifically for multiple drone failures in drone light shows, aimed at mitigating the risk of cascading collisions by drone evacuation and enabling rapid recovery from failures by leveraging strategically placed hidden drones. Our algorithm integrates a Social LSTM model with attention mechanisms to predict the trajectories of failing drones and compute near-optimal evacuation paths that minimize the likelihood of surviving drones being hit by fallen drones. In the recovery node, our system deploys hidden drones (operating with their LED lights turned off) to replace failed drones so that the drone light show can continue. Our experiments showed that our approach can greatly increase the robustness of a multi-drone system by leveraging deep learning to predict the trajectories of fallen drones.

\end{abstract}



\sloppy

\section{Introduction}
\label{sec:introduction}

\setcounter{footnote}{0}

\begin{figure}[t]
    \centering
    \includegraphics[width=0.9\linewidth]{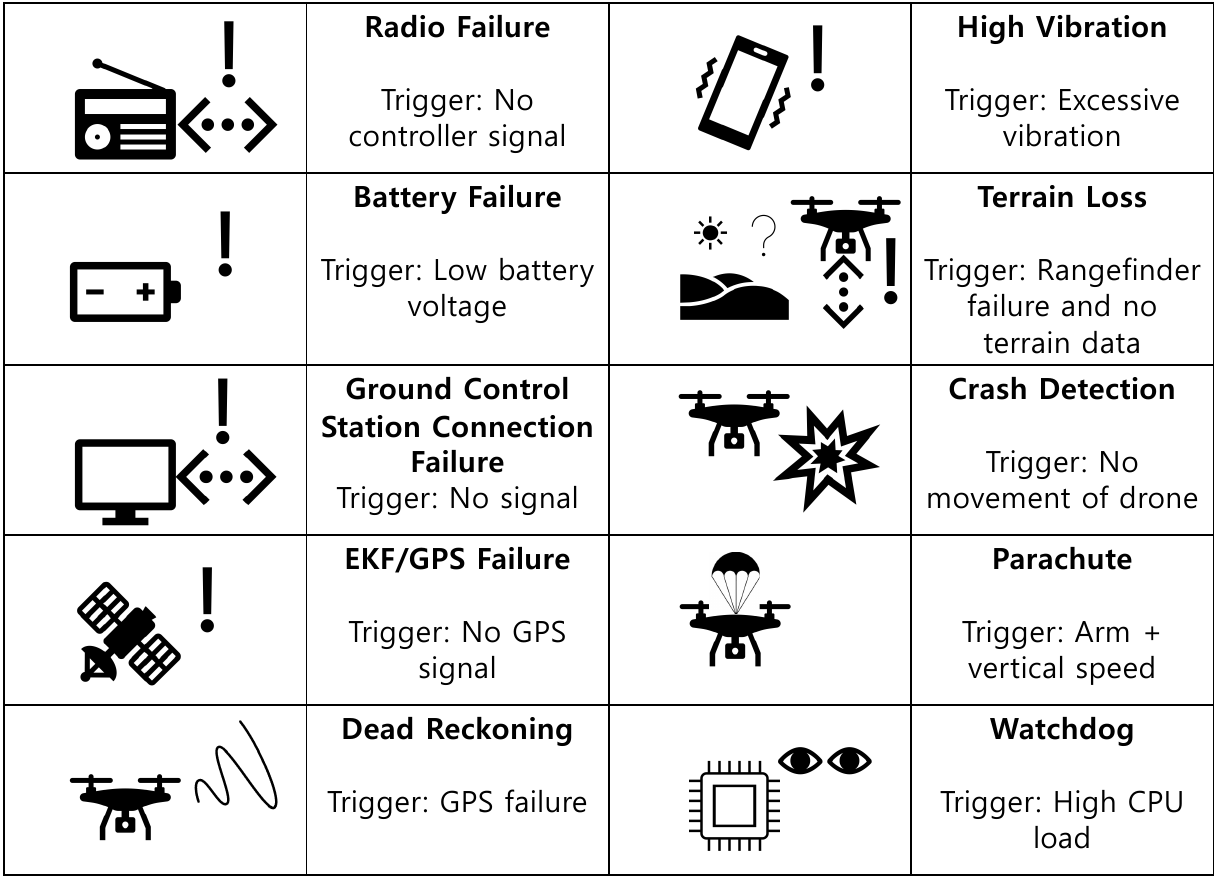}
    \caption{Ardupilot's failsafe triggers.}
    \label{fig:failure}

\mbox{}

    \includegraphics[width=0.9\linewidth]{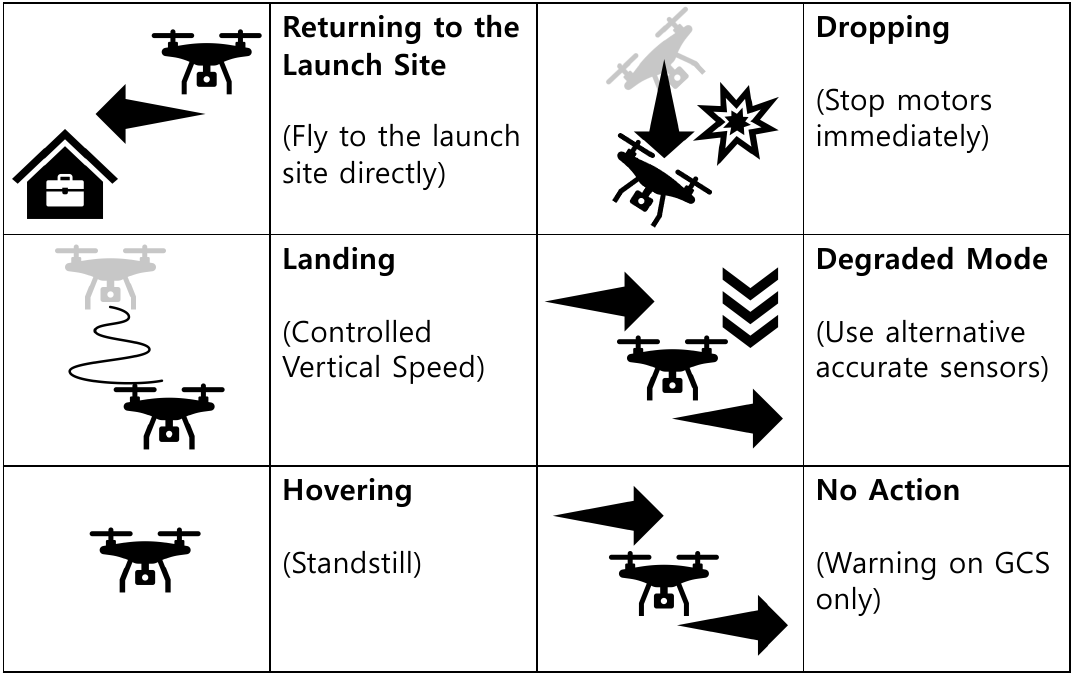}
    \caption{Ardupilot's failure responses.}
    \label{fig:response}
    \vspace{-.4cm}
\end{figure}

Drone light shows have increasingly supplanted traditional fireworks as a popular form of entertainment in cultural festivals and sporting events. However, drone light shows have been subject to several large-scale operational failures. Drone dropping incidents were documented in (1) Victoria Harbor in Hong Kong in 2018\footnote{\url{https://gpspatron.com/jamming-criminal}}, (2) Taichung City in Taiwan in 2020\footnote{\url{https://www.taipeitimes.com/News/taiwan/archives/2020/02/24/2003731529}}, (3) SeaTac in Washington in the United States in 2024\footnote{\url{https://www.seattletimes.com/seattle-news/seatacs-40000-fourth-of-july-fail-55-drones-drop-into-angle-lake}}, and (4) Ho Chi Minh City in Vietnam in 2025\footnote{\url{http://youtu.be/pgIuKu7kCLY}}. While some of these failures have been attributed to intentional interference, such as jamming of communication or GPS signals, others remain of indeterminate cause. In practice, drones are programmed with \emph{failure modes}, which initiate predefined maneuvers intended to mitigate damage to the drones. For example, Figs.~\ref{fig:failure} and \ref{fig:response} show the failsafe mechanisms and corresponding responses in Ardupilot, an open-source autopilot framework~\cite{meier2015ardupilot}. However, detecting the precise onset of a failure is often challenging. Even when a drone executes a programmed failsafe response, such as a controlled descent, the base station cannot reliably predict its trajectory or coordinate evasive actions by other drones. This problem is exacerbated by the dense spatial configuration of the drone swarms, where a falling drone can collide with nearby drones, triggering cascading failures. 

To address these challenges, we propose a drone parking algorithm designed to evacuate drones during multi-drone failures in drone dropping incidents. This method relies on a discretization of space and time, referred to as a space-time grid. Within this framework, the algorithm estimates the probability that falling drones will impact specific grid cells and subsequently generates motion plans to reposition unaffected drones while minimizing their likelihood of collision. This optimization problem is formulated as finding a \emph{nearly-optimal} path in a space-filling random graph constructed using the rapidly exploring random tree (RRT) algorithm that minimizes the collision probability. To the best of our knowledge, we are the first to consider drone evacuation during drone dropping incidents in the literature.


\section{Related Work}
\label{sec:related}

Popular autopilot software for hobbyist drones typically provides programmable failsafe responses, such as automated landing in response to basic triggers like low battery voltage~\cite{betaflight_failsafe}. More advanced autopilots, including Ardupilot and PX4~\cite{ardupilot_failsafe,bib:Chiu:px4_paper}, offer sophisticated per-drone failure detection and mitigation mechanisms, such as GNSS position-loss handling, enabling greater autonomous operation. Specialized software for drone light shows, such as the open-source Skybrush platform, extends these capabilities to the swarm level by incorporating geofence enforcement, pre-flight trajectory visualization, inter-drone distance monitoring, and maximum velocity constraints~\cite{skybrush_paper}. While these approaches can prevent many mid-flight collisions, they increase operational complexity and provide limited mitigation strategies when multiple drones fail simultaneously.

Several studies have investigated methods for constructing fault-tolerant and robust multi-agent systems capable of recovering from individual drone failures. Kumar and Cohen proposed an Adaptive Agent Architecture that leverages a brokered framework and collaborative teamwork to recover from failures~\cite{bib:Chiu:kumar2000towards}. Karimadini and Lin examined fault-tolerant cooperative tasking strategies to ensure global task completion despite partial multi-agent system failures~\cite{karimadini2011fault}. Jafari et al. developed an algorithm for optimal leader selection in leader-follower systems to maintain controllability under agent loss~\cite{bib:Chiu:jafari2011leader}. Chen et al. introduced a leaderless distributed adaptive protocol that compensates for faults in individual agents~\cite{bib:Chiu:chen2014fault}. Zhou et al. proposed a resilience evaluation framework for unmanned autonomous swarms using the Couzin-Leader model, assessing system performance following partial swarm failures~\cite{zhou2024resilience}.


\section{Trajectory Prediction of Fallen Drones}
\label{sec:defn}

The drone light show is performed in a finite 3D airspace, including the platform on which drones take off or land. We call the airspace the \emph{stage}. We assume there is no obstacle, such as buildings or trees, on the stage. We subdivide the stage into a 3D \emph{grid}, which consists of equally-sized grid cells, each of which is a small cubic region. The \emph{stage space} $\StageSpace$ is the set of grid cells occupied by the stage. The space-time of the drone show can be subdivided into a set of \emph{tiles}, each of which is a pair $\tile = (\cell,t)$, where $\cell$ is a grid cell and $t$ is a time step. The \emph{stage zone} is $\StageZone = \{ (\cell, t) \}_{\cell \in \StageSpace, \FirstTime \leq t < \LastTime}$, which is the set of all tiles between $\FirstTime$ and $\LastTime$, which are the start time and the end time of the show, respectively. In this paper, the word ``zone'' refers to a set of tiles, whereas the word ``space'' refers to a set of grid cells. The \emph{footprint} $\TileSet(\drone, t)$ of a drone $\drone$ at time $t$ is the set of tiles whose cells are \emph{occupied} by $\drone$ at $t$. That is, $\TileSet(\drone, t)$ includes any tiles that overlap with the body of $\drone$ at $t$. We say two drones $\drone_1$ and $\drone_2$ \emph{collide} with each other at time $t$ if and only if $\TileSet(\drone_1, t) \cap \TileSet(\drone_2, t) \neq \emptyset$. The \emph{footprint} $\TileSet(\plan, \drone, t)$ of a plan $\plan$ executed by $\drone$ at time $t$ is $\TileSet(\plan, \drone, t) = \bigcup_{t \leq t' \leq t + |\plan|} \TileSet(\drone, t')$, which includes all tiles occupied by $\drone$ when flying on the trajectory $\traj[\plan]$ between $t$ and $t + |\plan|$. When talking about the footprint of a plan $\plan$, we assume we know the drone that executes $\plan$ and the time of the execution from the context. To simplify our discussion, we will simply denote the footprint of $\plan$ by $\TileSet(\plan)$. We say two plans $\plan_1$ and $\plan_2$ \emph{collide} with each other if and only if $ \TileSet(\plan_1) \cap \TileSet(\plan_2) \neq \emptyset$. A formation plan $\PlanSet$ is \emph{collision-free} if every pair of plans in $\PlanSet$ does not collide with each other.

We trained a neural network to predict the trajectory of a fallen drone based on its poses in several time steps preceding $\StartTime$, the time when the incident occurs. Social LSTM~\cite{bib:Alahi16Social} is among the earliest models designed to forecast future trajectories of individuals in crowded environments using their historical positions. Vemula et al.~\cite{bib:Vemula18Social} enhanced Social LSTM with an attention mechanism to capture the relative influence of neighboring agents. Giuliari et al.~\cite{bib:Giuliari20Transformer} replaced LSTMs with transformer networks for trajectory forecasting. In our experiments, we employed a Social LSTM model augmented with the attention mechanism as described in~\cite{bib:Vemula18Social}.

The training data were generated using a simulator developed in-house, with real drone data employed to calibrate the drones’ behavior. We implemented three failure modes—(1) dropping, (2) landing, and (3) returning to the launch site---as illustrated in Fig.~\ref{fig:response}. To generate the dataset, a drone was first instantiated and made to follow a trajectory representative of a typical drone light show. Next, one of the failure modes was randomly triggered during flight, and the resulting trajectory of the fallen drone was recorded. This process was repeated many times to collect a dataset of trajectories, each associated with a sequence of drone poses observed in the few time steps preceding the fall. We assume that the specific failure mode is unknown and therefore do not label the trajectories with failure mode information. Finally, a Social LSTM model was trained to predict the trajectory of a fallen drone based on the sequence of poses preceding the failure, regardless of which of the three failure modes occurred.

\section{Occupancy Probability}
\label{sec:occupy}

The \emph{occupancy probability} $\Prob(\tile)$ of a tile $\tile$ is the probability that $\tile$ is occupied by any drone. The concept of occupancy probability is similar to the occupancy grid for mapping~\cite{bib:Thrun96Integrating,bib:Thrun05Probabilistic}. When there is no drone dropping incident, the occupancy probability can be derived from $\formation_{\sf first}$ and the original formation plan $\PlanSet^{\sf assigned}$ as follows: $\Prob(\tile) = 1$ for all $\tile \in \bigcup_{\plan^{\sf assigned} \in \PlanSet^{\sf assigned}} \TileSet(\plan^{\sf assigned})$ and $\Prob(\tile) = 0$ for all other tiles in $\StageZone$. When a drone dropping incident occurs, $\Prob(\tile)$ for any tile with time after $\StartTime$ may change since (1) the fallen drones in $\FallenDroneSet$ will no longer follow their assigned plans, and (2) some active drones in $\ActiveDroneSet$ can no longer follow their assigned plans in $\PlanSet^{\sf assigned}$.

Given a fallen drone $\drone \in \FallenDroneSet$, we utilize the machine learning model to estimate the occupancy probability of the tiles as follows. Initially, we set the occupancy probability of any tile after $\StartTime$ to zero, except the tiles in the footprints of the assigned trajectory of $\drone$ on or before $\StartTime$, which have the occupancy probability $1$. We perform $N$ random rollouts of the model to generate $N$ trajectories starting at $\StartTime$, where $N$ is a large positive integer, given the poses before $\StartTime$. For any tile $\tile$ whose time is greater than $\StartTime$, we count how many times $\tile$ is occupied by any trajectories. Then the occupancy probability of $\tile$ is $\Prob_{\drone}(\tile) = C / N$, where $C$ is the count.

\begin{figure}[t]
    \centering
    \includegraphics[width=0.95\linewidth]{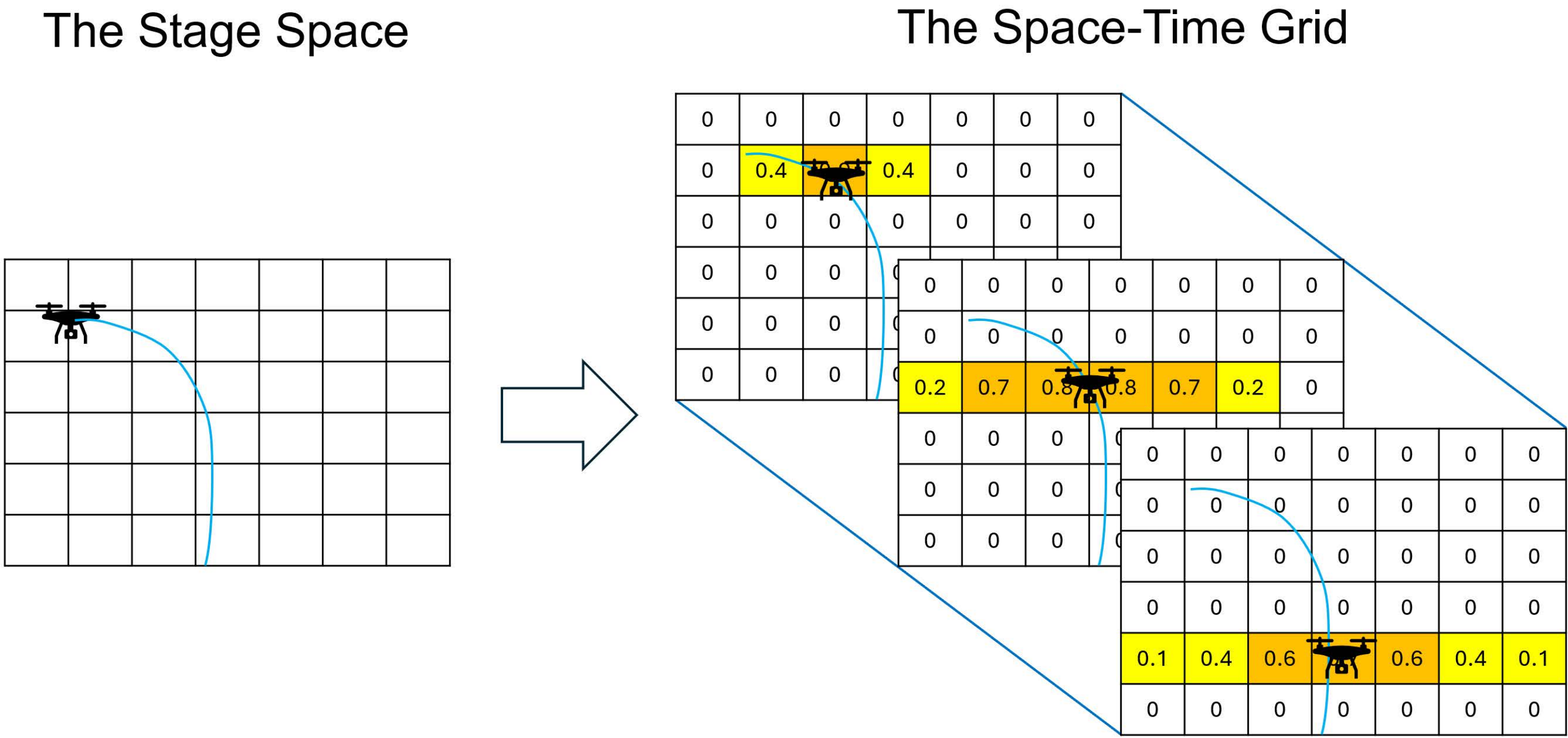}
    \caption{The occupancy probability in the fall zone of a drone.}
    \label{fig:occupancy}
    \vspace{-.5cm}
\end{figure}

The \emph{fall zone} $\FallZone(\drone)$ of a fallen drone $\drone$ is the set of tiles whose occupancy probability in $\Prob_{\drone}$ is non-zero. Fig.~\ref{fig:occupancy} shows the occupancy probability in the fall zone of a drone. Let $\EndTime^{\drone}$ be the largest time among all tiles with non-zero occupancy probability in $\Prob_{\drone}$. Then the \emph{end time} of the incident is $\EndTime = \max_{\drone \in \FallenDroneSet} \EndTime^{\drone}$, which is the largest time at which any fallen drone occupies any tile. The \emph{hit zone} for $\FallenDroneSet$ is $\HitZone = \bigcup_{\drone \in \FallenDroneSet} \FallZone(\drone)$, which is the set of tiles in the fall zones of the fallen drones. When our system is notified that a drone dropping incident occurs at time $\StartTime$, it will (1) \emph{evacuate} all drones in the hit zone at time $\StartTime$ by moving them out of the hit zone and parking them at some designated locations called \emph{parking space}, and (2) prevent the remaining drones from entering the hit zone by \emph{detouring} them to the parking space. The \emph{safe zone} is $\SafeZone = \StageZone \setminus \HitZone$, which is the set of tiles not in the hit zone. The \emph{parking space} $\ParkSpace$ is defined as the set of grid cells such that the tiles with these grid cells are always in the safe zone $\SafeZone$.

\section{The Parking Problem}

When a drone dropping incident occurs, the drones in the hit zone have to migrate to the parking space $\ParkSpace$ and hover inside $\ParkSpace$ until the end time $\EndTime$ of the incident.  We define the parking problem as finding a \emph{valid} formation plan for $\EvacuateDroneSet$, which is the set of drones that need evacuation:
\begin{defn}
    \label{def:validpark}
    A collision-free formation plan $\PlanSet$ for $\EvacuateDroneSet$ at time $\StartTime$ is \emph{valid} if and only if for all $\drone \in \DroneSet$,
    all cells of the footprint of $\drone$ at time $\EndTime$ are in $\ParkSpace$ when executing $\PlanSet$ at $\StartTime$ (i.e., $\{ c \}_{ (c, t) \in \TileSet(\drone,\EndTime)} \subseteq \ParkSpace$ ). 
\end{defn}

Collisions may occur when the drones in $\EvacuateDroneSet$ migrate to the parking space. Definition~\ref{def:validpark}, however, does not concern itself with the collisions with the fallen drones. Therefore, we need to define the parking problem as finding a valid formation plan while minimizing the probability of collisions. First, we combine the occupancy probablity of $\Prob_{\drone}$ of all fallen drones into one probability mass function: $\Prob(\tile) = 1 - \prod_{\drone \in \FallenDroneSet} (1 - \Prob_{\drone}(\tile))$. Hence, $\Prob(\tile)$ is the combined occupancy probability of $\tile$ such that any hidden drones could occupy $\tile$. Second, we define the \emph{collision-free} probability of $\tile$ by $\ProbSafe(\tile) = 1 - \Prob(\tile)$. More generally, we define the collision-free probability of a set of tiles:
\[
  \ProbSafe(\TileSet) = \prod_{\tile \in \TileSet} \ProbSafe(\tile).
\]
Given the footprint $\TileSet(\pi)$ of a plan $\plan$, $\ProbSafe(\TileSet(\pi))$ is the probability that the drone will \emph{not} collide with any fallen drone when executing $\pi$. An optimization version of the parking problem is defined as finding a valid formation plan $\PlanSet$ such that the weighted sum of the collision-free probability of the plans in $\PlanSet$ is maximized.

\begin{defn}
    \label{def:optimalpark}
    A valid formation plan $\PlanSet$ is \emph{optimal} w.r.t. the collision-free probability $\ProbSafe$ if and only if the score
    \[
       \score(\PlanSet) = \sum_{ \drone_i \in \EvacuateDroneSet} \alpha_i \ProbSafe(\TileSet(\plan_i))
    \]
    is maximized, where (1) $\plan_i \in \PlanSet$ is the plan for $\drone_i$, and (2) $\alpha_i \in \mathbb{R}^+$ is the weight of the collision-free probability of $\drone_i$.    
\end{defn}


\section{The Parking Algorithm}
\label{sec:algm}

The parking algorithm generates a near-optimal, valid formation plan $\PlanSet$ that maximizes the score function $\score(\PlanSet)$. The key idea is to generate the trajectories of individual drones independently and subsequently integrate these plans into a collision-free formation using an existing collision-avoidance strategy. For each drone $\drone$ in $\EvacuateDroneSet$, the algorithm constructs a random graph connecting the drone to the designated parking area and computes the shortest path to a target location within that space. The random graph is a directed acyclic graph, where vertices represent spatial positions and directed edges correspond to feasible actions that move $\drone$ from one vertex to another within a single time step. The weight of a directed edge $(\pos_1, \pos_2)$ associated with an action $\action$ is defined as
\[
  W(\pos_1, \pos_2) = - \log (\ProbSafe(\TileSet_{\pos_1, \pos_2})),
\]
where $\TileSet_{\pos_1, \pos_2}$ denotes the set of tiles occupied by $\drone$ when executing a single-action plan $\langle \action \rangle$ from $\pos_1$ to $\pos_2$. The random graph $(V, E)$ is generated using the rapidly exploring random tree (RRT) algorithm~\cite{bib:LaValle98RRT}, originating from the drone's initial position $\pos_0$ at time $\StartTime$. The resulting graph is space-filling and progressively spans the entire stage, ensuring that some vertices reach the designated parking area~\cite{bib:LaValle98RRT, bib:Karaman11Sampling}. Once constructed, a single-source shortest path algorithm suitable for graphs with negative edge weights (e.g., Bellman-Ford) is applied to compute a shortest-path tree from $\pos_0$. Given that the random graph is a directed acyclic graph, an alternative approach using topological sorting with single-pass relaxation can also efficiently compute the shortest paths. Among all vertices corresponding to positions where $\drone$ can be accommodated within the parking space, the algorithm selects the vertex $\pos^*$ associated with the shortest path $E^*$ from $\pos_0$. The sequence of actions along $E^*$, from $\pos_0$ to $\pos^*$, is then concatenated to form the final plan $\plan^*$. The total cost of this shortest path is
\[
  C^* = \sum_{(\pos_1, \pos_2) \in E^*} W(\pos_1, \pos_2).
\]
Hence, $C^* = - \sum_{(\pos_1, \pos_2) \in E^*} \log (\ProbSafe(\TileSet_{\pos_1, \pos_2})) = - \log ( \prod_{(\pos_1, \pos_2) \in E^*} \ProbSafe(\TileSet_{\pos_1, \pos_2}) ) = - \log( \ProbSafe(\TileSet_{E^*}))$, where $\TileSet_{E^*}$ is the set of all tiles $\drone$ occupies when executing $\plan^*$. Consequently, the collision-free probability of $\drone$ traversing the path $E^*$ is given by $\ProbSafe(\TileSet_{E^*}) = e^{-C^*}$. Because $C^*$ represents the minimum cost among all paths to the parking space within the random graph, $\ProbSafe(\TileSet_{E^*})$ corresponds to the \emph{maximum} collision-free probability for $\drone$ reaching the designated parking location.


\section{Recovery Mode}
\label{sec:recovery}

After time $\EndTime$, all fallen drones have either landed or exited the performance area, and the system transitions into recovery mode to allow the light show to proceed. Because all parked drones are hidden, the recovery mode selects a subset of hidden drones $\HiddenDroneSet' \subseteq \HiddenDroneSet$ positioned near locations where active drones are missing. The system then computes a recovery formation plan $\PlanSet_{\sf recovery}$ to reposition the drones in $\HiddenDroneSet'$ to these vacant locations. Upon executing $\PlanSet_{\sf recovery}$, the drones in $\HiddenDroneSet'$ become active by illuminating their LEDs and assume the assigned trajectories of the missing drones, thereby restoring the drone light show.


\section{Summary and Future Work}
\label{sec:summary}

Robustness is an important topic in embodied AI systems. Our drone parking system achieves robustness in AI-based multi-drone systems by mitigating the risks associated with multi-drone failures, which have emerged as a significant threat to the reliability of drone light shows. Our system leverages a deep learning model to estimate the occupancy probability of grid cells within a space-time grid, identifies high-risk zones requiring evacuation, and computes motion plans that maximize collision-free probability. Our preliminary experimental results showed that this approach outperforms baseline strategies, and increases the chance of recovery from multi-drone failures. We can improve our system by refining the trajectory prediction in the future.


\section*{ACKNOWLEDGMENTS}

This work was supported by National Research Foundation of Korea RS-2022-NR069751 (or 2022R1A2C101216813).

\bibliography{bib/auto,bib/minhyuk,bib/minhyuk2,bib/chiu}

@string{aaai = {Proceedings of the AAAI Conference on Artificial Intelligence (AAAI)}}

@string{cvpr = {IEEE/CVF Conference on Computer Vision and Pattern Recognition}}

@string{icra = {IEEE International Conference on Robotics and Automation (ICRA)}}

@conference{bib:Giuliari20Transformer,
	author = {Giuliari, Francesco and Hasan, Irtiza and Cristani, Marco and Galasso, Fabio},
	booktitle = {International Conference on Pattern Recognition (ICPR)},
	my-key = {Giuliari20Transformer},
	pages = {10335--10342},
	title = {Transformer Networks for Trajectory Forecasting},
	year = {2020}}

@conference{bib:Vemula18Social,
	author = {Anirudh Vemula and Katharina Muelling and Jean Oh},
	booktitle = icra,
	my-key = {Vemula18Social},
	pages = {4601--4607},
	title = {Social Attention: Modeling Attention in Human Crowds},
	year = {2018}}

@conference{bib:Alahi16Social,
	author = {Alahi, Alexandre and Goel, Kratarth and Ramanathan, Vignesh and Robicquet, Alexandre and Fei-Fei, Li and Savarese, Silvio},
	booktitle = {Proceedings of the IEEE Conference on Computer Vision and Pattern Recognition (CVPR)},
	my-key = {Alahi16Social},
	title = {Social {LSTM}: Human Trajectory Prediction in Crowded Spaces},
	year = {2016}}

@conference{bib:Thrun96Integrating,
	author = {Sebastian Thrun and Arno B\"{u}cken},
	booktitle = aaai,
	my-key = {Thrun96Integrating},
	pages = {944--950},
	title = {Integrating Grid-Based and Topological Maps for Mobile Robot Navigation},
	year = {1996}}

@article{bib:Karaman11Sampling,
	author = {Sertac Karaman and Emilio Frazzoli},
	journal = {The International Journal of Robotics Research},
	my-key = {Karaman11Sampling},
	number = {7},
	pages = {846--894},
	title = {Sampling-based algorithms for optimal motion planning},
	volume = {30},
	year = {2011}}

@book{bib:Thrun05Probabilistic,
	author = {Sebastian Thrun and Wolfram Burgard and Dieter Fox},
	my-key = {Thrun05Probabilistic},
	publisher = {The {MIT} Press},
	title = {Probabilistic Robotics},
	year = {2005}}

@techreport{bib:LaValle98RRT,
	author = {Steven M. LaValle},
	institution = {Computer Science Dept, Iowa State University},
	my-key = {LaValle98RRT},
	number = {TR 98-11},
	title = {Rapidly-Exploring Random Trees: A New Tool for Path Planning},
	year = {1998}}

@inproceedings{bib:Chiu:jafari2011leader,
  title={Leader selection in multi-agent systems subject to partial failure},
  author={Jafari, Saeid and Ajorlou, Amir and Aghdam, Amir G},
  booktitle={American Control Conference},
  pages={5330--5335},
  year={2011},
}

@inproceedings{bib:Chiu:kumar2000towards,
  title={Towards a fault-tolerant multi-agent system architecture},
  author={Kumar, Sanjeev and Cohen, Philip R},
  booktitle={International Conference on Autonomous Agents},
  pages={459--466},
  year={2000}
}

@article{bib:Chiu:chen2014fault,
  title={Fault-tolerant consensus of multi-agent system with distributed adaptive protocol},
  author={Chen, Shun and Ho, Daniel WC and Li, Lulu and Liu, Ming},
  journal={IEEE Transactions on Cybernetics},
  volume={45},
  number={10},
  pages={2142--2155},
  year={2014},
}

@INPROCEEDINGS{bib:Chiu:px4_paper,
  author={Meier, Lorenz and Honegger, Dominik and Pollefeys, Marc},
  booktitle={IEEE International Conference on Robotics and Automation (ICRA)},
  title={PX4: A node-based multithreaded open source robotics framework for deeply embedded platforms},
  year={2015},
  pages={6235-6240},
}

@string{aaai={AAAI}}

@string{icra={ICRA}}

@misc{betaflight_failsafe,
  author       = {{Betaflight Team}},
  title        = {Failsafe},
  howpublished = {Betaflight Wiki},
  year         = {2025},
  note         = {Date of Access: September 15, 2025},
  url          = {https://betaflight.com/docs/wiki/archive/Failsafe}
}

@misc{ardupilot_failsafe,
  author       = {{Ardupilot Team}},
  title        = {Failsafes},
  howpublished = {ArduPilot Copter documentation},
  year         = {2025},
  note         = {Date of Access: September 15, 2025},
  url          = {https://ardupilot.org/copter/docs/failsafe-landing-page.html}
}

@misc{skybrush_paper,
  author       = {Skybursh Team},
  title        = {Skybrush Studio},
  howpublished = {Skybrush Studio Description Webpage},
  year         = {2025},
  note         = {Date of Access: September 15, 2025},
  url          = {https://skybrush.io/modules/studio/}
}

@article{karimadini2011fault,
  title={Fault-tolerant cooperative tasking for multi-agent systems},
  author={Karimadini, Mohammad and Lin, Hai},
  journal={International Journal of Control},
  volume={84},
  number={12},
  pages={2092--2107},
  year={2011},
  publisher={Taylor \& Francis}
}

@article{zhou2024resilience,
  title={The resilience evaluation of unmanned autonomous swarm with informed agents under partial failure},
  author={Zhou, Xinxin and Huang, Yun and Bai, Guanghan and Xu, Bei and Tao, Junyong},
  journal={Reliability Engineering \& System Safety},
  volume={244},
  pages={109920},
  year={2024},
  publisher={Elsevier}
}

@inproceedings{meier2015ardupilot,
  title={Ardupilot: A flexible open source software for autonomous vehicles},
  author={Meier, Lorenz and STools, PX4 Development Team and others},
  booktitle={32nd Chaos Communication Congress (32C3)},
  year={2015},
}


\end{document}